\documentclass[conference]{IEEEtran}
\IEEEoverridecommandlockouts
\usepackage{multirow}
\usepackage{booktabs}
\usepackage{cite}
\usepackage{amsmath,amssymb,amsfonts}
\usepackage{algorithmic}
\usepackage{graphicx}
\usepackage{textcomp}
\usepackage{xcolor}
\usepackage{balance}
\def\BibTeX{{\rm B\kern-.05em{\sc i\kern-.025em b}\kern-.08em
    T\kern-.1667em\lower.7ex\hbox{E}\kern-.125emX}}
\begin{document}

\title{Localizing Emergent Failures in Agentic AI: Recovering Minimal Repair Families via Counterfactual Replay\\
% {\footnotesize \textsuperscript{*}Note: Sub-titles are not captured for https://ieeexplore.ieee.org  and
% should not be used}
% \thanks{Identify applicable funding agency here. If none, delete this.}
}

\author{
\IEEEauthorblockN{
Bingjie Li\IEEEauthorrefmark{1},
Yumeng Song\IEEEauthorrefmark{2},
Zhongming Yao\IEEEauthorrefmark{3},
and Tianyi Li\IEEEauthorrefmark{2}
}

\IEEEauthorblockA{
\IEEEauthorrefmark{1}
School of Computer Science and Engineering,
Northeastern University, Shenyang, China\\
libingjie@mails.neu.edu.cn
}

\IEEEauthorblockA{
\IEEEauthorrefmark{2}
Department of Computer Science,
Aalborg University, Aalborg, Denmark\\
\{yumengs,tianyi\}@cs.aau.dk
}

\IEEEauthorblockA{
\IEEEauthorrefmark{3}
College of Computer Science and Technology,
Zhejiang University, Hangzhou, China\\
yaozzzm@gmail.com
}
}

\maketitle

\begin{abstract}
Failures in agentic AI systems can arise from interactions among messages
exchanged by multiple large language model (LLM) agents. Pointwise attribution
cannot distinguish a jointly necessary repair from alternative singleton
repairs. We formulate Minimal Repair Family Recovery (MRFR): recovering all
inclusion-minimal event sets whose counterfactual replay restores task success
within a declared size bound. We propose Graph-Constrained Joint Replay
(GCJR), which slices failure-relevant events from an execution dependency
graph, constructs graph-feasible singleton and pair candidates, and verifies
them by replay with paired clean counterparts. For fixed replay outcomes,
GCJR is exact within its declared graph domain. On 90 in-scope cases from a
120-DAG controlled benchmark, GCJR achieves 1.000 Family Exact Match while
reducing mean replay calls from 56.3 to 25.3 (55.1\%) relative to exhaustive
search. On a 24-case, four-agent LLM pilot, it again achieves 1.000 Family
Exact Match and reduces mean model calls from 21.0 to 10.0 (52.4\%);
single-event replay misses jointly necessary repairs.
\end{abstract}

\begin{IEEEkeywords}
LLM multi-agent systems, failure localization, counterfactual replay, minimal
repair family, event dependency graph
\end{IEEEkeywords}

\section{Introduction}
\label{sec:introduction}

Agentic AI systems composed of multiple large language model (LLM) agents
solve complex tasks by assigning specialized roles and exchanging intermediate
results. Frameworks such as AutoGen, MetaGPT, and ChatDev illustrate the benefits of role-based collaboration~\cite{wu2023autogen,hong2024metagpt,qian2024chatdev}, while MultiAgentBench shows that communication protocols and interaction topologies materially affect task outcomes~\cite{zhu2025multiagentbench}. This distributed interaction structure also complicates debugging: an
incorrect final answer does not reveal which earlier interaction must change,
because an error may originate in a single message, propagate through
downstream agents, or arise only through the joint effect of multiple messages.

Existing diagnostic work has catalogued multi-agent failure modes~\cite{cemri2026multi} and localized failures to responsible agents or decisive steps~\cite{zhang2025agent,chen2026seeing}. Graph-based methods use information dependencies, hierarchical causal structure, or paired successful traces to narrow root-step search~\cite{zhang2025graphtracer,wang2026flat,zhu2026hpfa}. Intervention-based methods validate edited trajectories~\cite{ma2026dover,zhu2026raffles}, construct step-level counterfactual repairs~\cite{bonagiri2026causalflow}, or distribute responsibility across interacting steps~\cite{shah2026causal}. These advances make localization increasingly actionable, but leave a set-level question under-specified: which smallest combinations of interactions can restore success? A pointwise label or ranking cannot distinguish a jointly necessary failure, where two messages must both be repaired, from an alternative-repair failure, where either message is sufficient.

We study this question with \emph{Graph-Constrained Joint Replay} (GCJR). GCJR represents an execution trace as a directed interaction graph, replaces selected candidate messages with matched clean counterparts, and validates each intervention by replaying the workflow from an available checkpoint. It tests graph-feasible message sets in increasing cardinality and retains every inclusion-minimal successful intervention within a declared search bound. For convenience, we call this collection the \emph{minimal repair family}. Unlike a scalar responsibility score, the returned family can express singleton faults, jointly necessary repairs, and multiple alternative repairs. The method is deliberately oracle-assisted: it localizes failures under controlled replay but does not generate repair messages automatically.

Our contributions are threefold:
\begin{itemize}
    \item We expose a blind spot of pointwise failure attribution by separating joint necessity from alternative sufficiency, and define a replay-verifiable, set-level target over all inclusion-minimal successful interventions within an explicit search bound.
    \item We develop GCJR, a dependency-aware procedure that searches singleton and pair interventions, uses replay as a success certificate, and returns the complete family within its declared graph-feasible domain without collapsing it into per-event scores.
    \item  We provide controlled evidence on both symbolic workflows and a real
    LLM multi-agent system. GCJR matches exhaustive search in repair-family
    recovery while reducing replay calls by 52--55\%, and identifies jointly
    necessary repairs missed by pointwise baselines.
\end{itemize}

\section{Problem Formulation and GCJR}
\label{sec:method}

We first define the set-valued diagnostic target, then present
Graph-Constrained Joint Replay (GCJR) for recovering it, and finally state the
method's exactness and replay cost.

\subsection{Problem Formulation}
\label{sec:formulation}

We represent a failed multi-agent execution as a directed acyclic graph
$G=(V,E)$, where each node is an observable message or action and each edge
encodes an execution dependency. Let $a(v)$ denote the agent responsible for
event $v$, let $o$ be the observed failure sink, and let $C\subseteq V$ be the
declared set of intervenable events.

For an intervention set $S\subseteq C$, the replay backend replaces the
corresponding events with their paired clean messages, or equivalently removes
their injected corruptions, and reruns the downstream workflow. Let
$Y_j(S)\in\{0,1\}$ be the outcome of repeat $j$, where one denotes task success.
Using common replay seeds across interventions, we define
\begin{equation}
 \widehat p(S)=\frac{1}{r}\sum_{j=1}^{r}Y_j(S), \qquad
 h_\theta(S)=\mathbb I[\widehat p(S)\geq\theta].
 \label{eq:replay-success}
\end{equation}
We assume $h_\theta(\varnothing)=0$, i.e., the unmodified execution remains a
failure under the replay protocol.

\begin{figure}[htbp]
    \centering
    \includegraphics[width=0.75\linewidth]{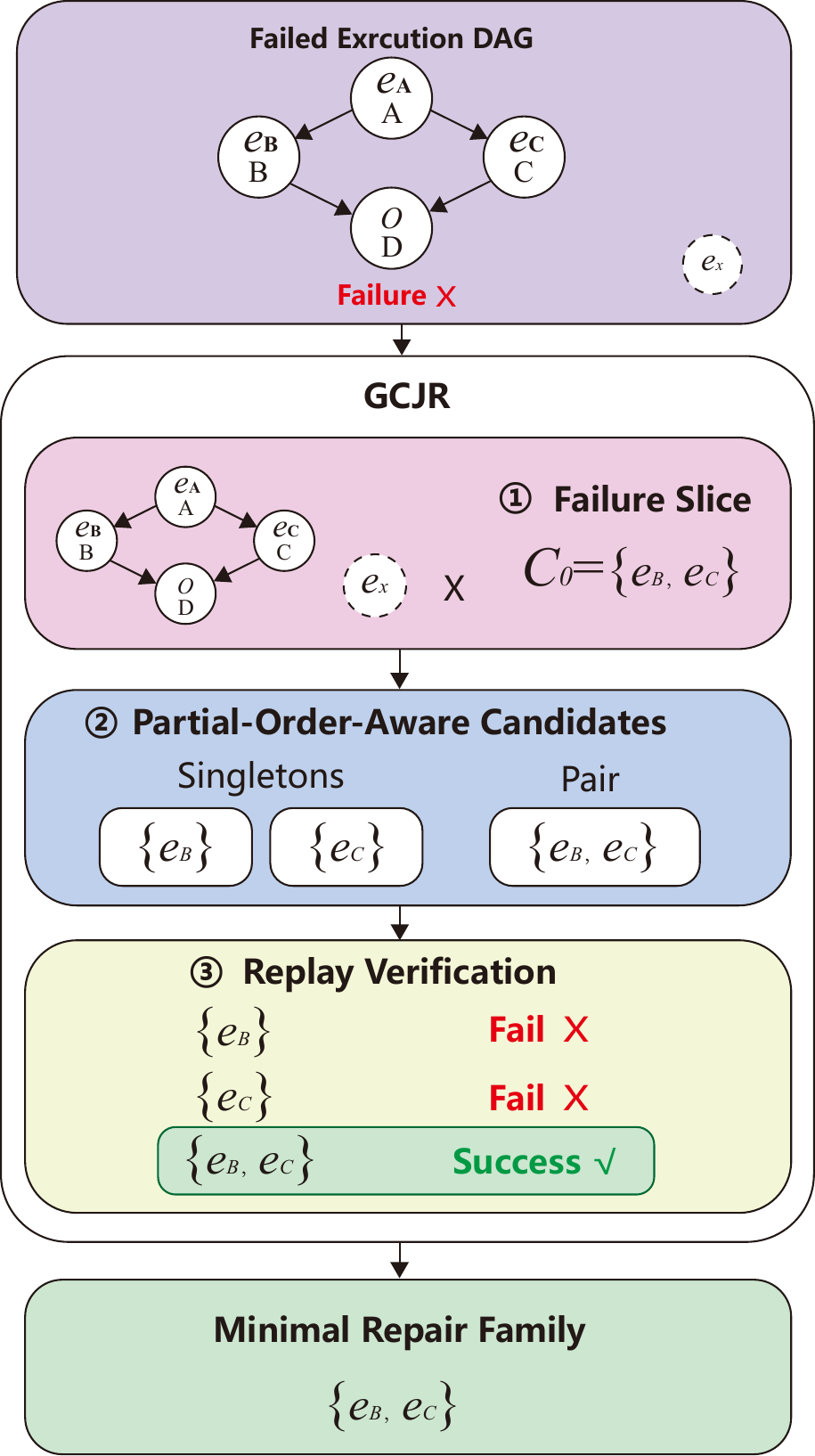}
    \caption{Overview of the proposed framework.}
    \label{fig:framework}
\end{figure}
Given a maximum repair size $q$, our target is the family of all successful
interventions whose strict subsets fail:
\begin{equation}
\begin{aligned}
\mathcal F_{q,\theta}=\bigl\{S\subseteq C:\;&1\leq|S|\leq q,\\[-1mm]
&h_\theta(S)=1,\quad
h_\theta(T)=0\ \forall T\subsetneq S\bigr\}.
\end{aligned}
\label{eq:minimal-family}
\end{equation}
Minimality is defined by set inclusion. Hence $\{\{u,v\}\}$ denotes a jointly
necessary repair, whereas $\{\{u\},\{v\}\}$ denotes two alternative repairs.

\subsection{Graph-Constrained Joint Replay}
\label{sec:gcjr}

GCJR recovers the target through three stages: failure slicing, graph-constrained
candidate construction, and replay verification. We study singleton and pair
repairs, i.e., $q\leq2$.

\paragraph{Failure slicing.}
GCJR first removes events that cannot structurally affect the observed failure.
It restricts intervention candidates to the backward slice of $o$:
\begin{equation}
 B(o)=\operatorname{Anc}_{G}(o)\cup\{o\}, \qquad C_o=C\cap B(o).
 \label{eq:backward-slice}
\end{equation}
This screening is sound when the dependency graph contains every influence path
relevant to the replay outcome.

\paragraph{Candidate construction.}
Write $u\parallel_G v$ when neither event reaches the other. Besides all
singletons in $C_o$, GCJR admits a pair only if its events come from different
agents, are causally incomparable, and meet at a downstream join:
\begin{equation}
\begin{aligned}
\mathcal P_G
&=\bigl\{\{u,v\}\subseteq C_o:a(u)\neq a(v),\\[-1mm]
&\quad u\parallel_G v,\ \exists z\in B(o):
u,v\leadsto z,\ \deg_G^-(z)\geq2\bigr\},\\
\mathcal D_G
&=\{\{v\}:v\in C_o\}\cup\mathcal P_G.
\end{aligned}
\label{eq:gcjr-domain}
\end{equation}
The graph thus proposes a bounded candidate domain $\mathcal D_G$; it does not
by itself certify that any candidate repairs the task.

\paragraph{Replay verification and family recovery.}
GCJR evaluates $\mathcal D_G$ in increasing cardinality. After confirming the
empty intervention fails, it replays every singleton in $C_o$ and retains each
successful singleton. Any pair containing such a singleton is skipped because
it cannot be inclusion-minimal. GCJR then replays every remaining pair in
$\mathcal P_G$ and retains the successful ones. Requests are memoized, and the
same replay seeds are shared across interventions. Consequently, a returned
singleton $\{u\}$ satisfies $h_\theta(\varnothing)=0$ and
$h_\theta(\{u\})=1$, while a returned pair $\{u,v\}$ additionally satisfies
$h_\theta(\{u\})=h_\theta(\{v\})=0$ and $h_\theta(\{u,v\})=1$.

\subsection{Domain Exactness and Replay Cost}
\label{sec:method-analysis}

\paragraph{Domain exactness.}
For fixed replay outcomes, GCJR returns exactly the inclusion-minimal
successful sets in $\mathcal D_G$. It tests every singleton and skips an
admissible pair only when a successful singleton has already certified that
pair as non-minimal; every other admissible pair is tested. Therefore, if
$\mathcal F_{2,\theta}\subseteq\mathcal D_G$, GCJR recovers the complete
bounded repair family $\mathcal F_{2,\theta}$.

\paragraph{Replay cost.}
Let $n=|C|$, $m=|C_o|$, and $p$ be the number of admissible pairs remaining
after successful-singleton pruning. With $r$ repeats per intervention,
\begin{equation}
\begin{aligned}
Q_{\mathrm{GCJR}} &= r(1+m+p),
& p &\leq \binom{m}{2},\\
Q_{\mathrm{exh}} &= r\left(1+n+\binom{n}{2}\right).
\end{aligned}
\label{eq:replay-cost}
\end{equation}
The reduction comes from failure slicing, structural pair screening, and
minimality pruning.

\begin{table}[t]
\centering
\caption{Statistics of the experimental workloads.}
\label{tab:benchmark-summary}
\footnotesize
\setlength{\tabcolsep}{3pt}
\renewcommand{\arraystretch}{1.06}
\begin{tabular*}{\columnwidth}
{@{\extracolsep{\fill}}lcc@{}}
\toprule
Statistic & Controlled-DAG & DualSolve-MAS \\
\midrule
Cases      & 120       & 24 \\
Schedules  & 360       & -- \\
Agents     & 4         & 4 \\
Candidates & 6--14     & 3 \\
Families   & S/J/A/SJ  & S/J/A \\
Replay     & Simulator & Qwen ($r=3$) \\
\bottomrule
\end{tabular*}
\end{table}

\begin{table*}[t]
\centering
\caption{Family recovery and replay cost on the 90 in-scope Controlled-DAG cases.}
\label{tab:symbolic-results}
\small
\setlength{\tabcolsep}{5pt}
\renewcommand{\arraystretch}{1.08}
\begin{tabular*}{\textwidth}{@{\extracolsep{\fill}}l c c r c@{}}
\toprule
\multirow[c]{2}{*}{Method} & \multicolumn{2}{c}{Family recovery} & \multicolumn{2}{c}{Replay efficiency} \\
\cmidrule(lr){2-3}\cmidrule(lr){4-5}
& J FEM $\uparrow$ & Macro FEM $\uparrow$ & Calls $\downarrow$ & Matched saving $\uparrow$ \\
\midrule
Graph-blind exhaustive & 1.000 & \textbf{1.000} & 56.3 & 0.0\% \\
Single-event replay & 0.000 & 0.667 & 10.7 & N/A \\
Execution-window replay & 0.100 & 0.700 & 14.4 & N/A \\
Exact Shapley top-2 & 1.000 & 0.333 & 3285.3 & N/A \\
Centrality top-2 & 0.033 & 0.011 & 0.0 & N/A \\
Random pair & 0.011 & 0.004 & 0.0 & N/A \\
\midrule
\textbf{GCJR (ours)} & 1.000 & \textbf{1.000} & \textbf{25.3} & \textbf{55.1\%} \\
\bottomrule
\end{tabular*}
\end{table*}
\begin{table}[t]
\centering
\caption{Family recovery and LLM-call cost on the 24-case DualSolve-MAS pilot.}
\label{tab:real-agent-results}
\footnotesize
\setlength{\tabcolsep}{2.5pt}
\renewcommand{\arraystretch}{1.08}
\begin{tabular*}{\columnwidth}{@{\extracolsep{\fill}}l c c r c@{}}
\toprule
\multirow[c]{2}{*}{Method} & \multicolumn{2}{c}{Family recovery} & \multicolumn{2}{c}{LLM efficiency} \\
\cmidrule(lr){2-3}\cmidrule(lr){4-5}
& J FEM $\uparrow$ & Macro FEM $\uparrow$ & Calls $\downarrow$ & Saving $\uparrow$ \\
\midrule
Exhaustive & 1.000 & \textbf{1.000} & 21.0 & 0.0\% \\
Single-event & 0.000 & 0.667 & 12.0 & N/A \\
Execution-window & 0.000 & 0.667 & 13.0 & N/A \\
LLM log-only & 0.000 & 0.333 & 1.0 & N/A \\
\midrule
\textbf{GCJR (ours)} & 1.000 & \textbf{1.000} & \textbf{10.0} & \textbf{52.4\%} \\
\bottomrule
\end{tabular*}
\end{table}

% Requires \usepackage{booktabs} in the preamble.
\section{Experimental Evaluation}
\label{sec:experiments}

We evaluate whether GCJR recovers the complete minimal repair family and whether
graph constraints reduce replay cost without changing that family. Both
questions are tested in a symbolic benchmark and a real-agent pilot.

\subsection{Benchmarks and Protocol}

Table~\ref{tab:benchmark-summary} summarizes two controlled workloads. The
\emph{Controlled-DAG} suite instantiates a four-agent fork--join workflow
$A\rightarrow\{B,C\}\rightarrow D$ with integer operations and Boolean outcome
checks. It contains 30 logical DAGs for each of Single, Joint-AND,
Alternative-OR, and Sequential-Joint. The first three types form the 90-DAG
in-scope set; Sequential-Joint is an out-of-scope control because its two faulty
events are causally comparable.
Each DAG contains 6--14 candidates, including irrelevant distractors, and is
executed under three valid topological schedules.

\subsection{Main Results}

\emph{DualSolve-MAS} is a 24-case real-agent pilot with eight program-generated
arithmetic cases per in-scope type. Planner A, Solvers B/C, and Aggregator D are
instantiated by a local Qwen2.5-1.5B-Instruct model. An
exact arithmetic executor supplies correct tool results to the solvers so that
the experiment isolates communication localization rather than arithmetic
ability. We deterministically corrupt the numeric message sent by B and/or C;
an intervention restores the corresponding paired clean message and reruns D.
The third candidate is an irrelevant audit event.

Methods observe only the graph, candidate set, agent identities, and replay
outcomes, not fault labels or clean values. Controlled-DAG uses deterministic
replay with seed 20260826; confidence intervals use 2,000 DAG-level bootstrap
samples after aggregating the three schedules. DualSolve-MAS uses temperature
0, at most 80 new tokens, $r=3$, and success threshold $\theta=2/3$.

\subsection{Baselines and Metrics}

All replay methods share the same candidates and intervention backend.
\emph{Graph-blind exhaustive} enumerates every set with $|S|\leq2$ and is the
primary accuracy--cost comparator. \emph{Single-event replay} tests only
first-order interventions, while \emph{execution-window replay} uses proximity
in the linear log instead of partial-order constraints. Exact Shapley
top-2~\cite{shapley1953value}, betweenness-centrality top-2~\cite{freeman1978centrality},
and a seeded random pair produce fixed-cardinality rankings rather than repair
families. The real-agent pilot also includes a one-call LLM judge that predicts
a family from the failed log without intervention.

Family Exact Match (FEM) is one iff the predicted antichain equals the exhaustive
minimal-repair family. Antichain Jaccard (AJ) measures partial overlap, Repair
Success at 1 (RS@1) tests the first returned set, Minimality Violation (MV)
detects non-minimal outputs, and Schedule Consistency (SC) measures agreement
across symbolic linearizations. We report replay executions for Controlled-DAG
and real LLM calls for DualSolve-MAS in separate tables.

\subsection{Main Results}

GCJR recovers all 90 in-scope symbolic families, with AJ, RS@1, and SC of 1.000
and MV=0. Its mean replay cost is 25.3 (95\% CI: 23.1--27.5), versus 56.3
(51.0--61.9) for exhaustive search, a 55.1\% reduction. Single-event and
execution-window replay miss jointly necessary repairs. Exact Shapley top-2
attains RS@1=1.000 but only 0.333 macro FEM and 0.667 MV because it represents
Alternative-OR cases as a successful but non-minimal pair. Thus, a set-valued
target preserves distinctions lost by fixed-cardinality rankings.

In the real-agent pilot, GCJR matches the exhaustive family on all 24 cases,
with RS@1=1.000 and MV=0, while reducing mean LLM calls from 21.0 to 10.0
(52.4\%). Single-event, execution-window, and log-only methods obtain
Joint-AND FEM=0.

\subsection{Ablations and Boundary Tests}

Restricting GCJR to singleton interventions ($q=1$) reduces Joint-AND
FEM from 1.000 to 0.000, confirming that singleton-only replay cannot
recover jointly necessary repairs. Removing the minimality filter reduces
Alternative-OR FEM from 1.000 to 0.000 and increases MV from 0 to 0.783,
showing that successful replay alone is insufficient to recover the complete
minimal repair family. GCJR abstains on all 30 out-of-scope Sequential-Joint
controls, consistent with its restriction to causally incomparable pairs.
Across three sampled topological linearizations of each DAG, GCJR returns
identical repair families (SC=1.000), demonstrating invariance to the ordering
of causally independent events in our deterministic setting.

GCJR assumes that the observed dependency graph preserves the causal paths
relevant to the failure. When 10\% of its edges are randomly hidden from the
diagnostic method, in-scope macro FEM decreases from 1.000 to 0.552. Although
at least one ground-truth minimal repair set remains fully visible in the
failure slice for 73.3\% of the cases, missing edges can exclude valid
candidates during slicing or structural screening, preventing recovery of the
complete repair family.
\section{Related Work}
\label{sec:related}

\textbf{Failure analysis in multi-agent systems.}
Recent work has moved beyond end-task accuracy toward diagnosing why an
LLM-based multi-agent system fails. MAST organizes recurrent failure modes,
while Who\&When, Seeing the Whole Elephant, and SCOPE identify responsible
agents, steps, or components from execution traces
\cite{hu2023spatiotemporaltrajectorysimilaritymeasures,zhang2025agent,chen2026seeing,sun2026scope}. These methods
provide useful taxonomies or attribution signals. Our target is different: the
output is not one label or a ranking, but the complete family of
inclusion-minimal event sets within a declared size-bounded intervention domain.

\textbf{Graph-guided attribution and replay.}
GraphTracer, CHIEF, and HPFA show that structural or causal graphs can improve
failure attribution \cite{zhang2025graphtracer,wang2026flat,zhu2026hpfa}; thus,
using a graph is not itself our novelty. AgenTracer, DoVer, CausalFlow, and
Causal Agent Replay further use replay or counterfactual interventions to
identify decisive steps, validate diagnoses, or estimate responsibility
\cite{zhang2026agentracer,ma2026dover,bonagiri2026causalflow,shah2026causal}.
Most recently, SymTrace records execution prefixes and intervention anchors to
separate causal repair from stochastic resampling \cite{luan2026repairresamplerethinkingfailure}. These
systems primarily return a failure location, an intervention effect, or a
selected repair. In contrast, GCJR enumerates the complete bounded minimal
repair family within a declared graph domain. This set-valued output preserves
the distinction between jointly necessary events (AND) and alternative
singleton repairs (OR), which a single attribution ranking cannot represent.

\textbf{Minimal diagnosis and fault localization.}
Minimal diagnoses and causes have long been studied through consistency,
structural counterfactuals, delta debugging, and program repair
\cite{reiter1987theory,halpern2005causes,zeller2002simplifying,rothenberg2020must}.
Accordingly, we do not claim that minimal sets or counterfactual testing are
new in the abstract. Our contribution is a bounded instantiation for
multi-agent event graphs: replay-validated repair families, an explicit
antichain target, and graph-based pruning whose exactness is stated only for
the declared candidate domain.

\balance
\bibliography{reference}
\end{document}